\documentclass[10pt,twocolumn,letterpaper]{article}

\usepackage{iccv}
\usepackage{times}
\usepackage{epsfig}
\usepackage{graphicx}
\usepackage{amsmath}
\usepackage{amssymb}

\usepackage{booktabs}
\usepackage{pifont}
\usepackage[dvipsnames]{xcolor}
\usepackage{wrapfig}
\usepackage{dblfloatfix}
\usepackage{tablefootnote}
\usepackage{multirow}
\usepackage{enumitem}
\usepackage[symbol]{footmisc}

\usepackage{textcomp}
\newcommand{\textapprox}{\raisebox{0.5ex}{\texttildelow}}

\definecolor{orange}{HTML}{FF7F00}
\definecolor{col1}{HTML}{b4ce08}
\definecolor{col2}{HTML}{407934}
\definecolor{col3}{HTML}{353B3D}
\definecolor{airforceblue}{rgb}{0.36, 0.54, 0.66}
\definecolor{bleudefrance}{rgb}{0.19, 0.55, 0.91}
\definecolor{bluemunsell}{rgb}{0.0, 0.5, 0.69}
\definecolor{darkpastelgreen}{rgb}{0.01, 0.75, 0.24}
\definecolor{armygreen}{rgb}{0.29, 0.33, 0.13}
\definecolor{cadmiumgreen}{rgb}{0.0, 0.42, 0.24}
\definecolor{darkspringgreen}{rgb}{0.09, 0.45, 0.27}
\definecolor{ferngreen}{rgb}{0.31, 0.47, 0.26}
\definecolor{forestgreen(web)}{rgb}{0.13, 0.55, 0.13}
\definecolor{grannysmithapple}{rgb}{0.66, 0.89, 0.63}
\definecolor{green(html/cssgreen)}{rgb}{0.0, 0.5, 0.0}

\definecolor{col4}{HTML}{4B5232}

\newcommand{\textcite}{\cite}

\newcommand\textual[1]{{\color{bluemunsell}{#1}}}
\newcommand\visual[1]{$\color{orange} {\mathrm{#1}^{V}}$}
\newcommand\answer[1]{\textbf{\color{darkpastelgreen}{#1}}}

\newcommand{\vi}{v_i}
\newcommand{\qi}{q_i}
\newcommand{\ai}{a_i}

\newcommand{\positive}[1]{\scriptsize{\color{green(html/cssgreen)}{#1}}}
\newcommand{\negative}[1]{\scriptsize{\color{red}{#1}}}
\newcommand{\white}[1]{\scriptsize{\color{white}{#1}}}

\usepackage{array}
\newcolumntype{P}[1]{>{\centering\arraybackslash}p{#1}}

\usepackage[pagebackref=true,breaklinks=true,letterpaper=true,colorlinks,bookmarks=false]{hyperref}

\iccvfinalcopy 

\def\iccvPaperID{1960} 

\ificcvfinal\fi

\begin{document}

\title{Beyond Question-Based Biases:\\Assessing Multimodal Shortcut Learning in Visual Question Answering}

\author{
    Corentin Dancette$^1$\hspace{0.07cm}\footnotemark[1] \hspace{1cm} 
    Rémi Cadène$^{1,2}$ \footnotemark[1] \hspace{-0.05cm} \footnotemark[2] \hspace{1cm}
    Damien Teney$^{3,4}$ \hspace{1cm}
    Matthieu Cord$^{1,5}$
    \\
$^1$Sorbonne Université, CNRS, LIP6, 4 place Jussieu, Paris \\ 
$^2$Carney Institute for Brain Science, Brown University, USA \hspace{0.5cm} $^3$Idiap Research Institute \\
\hspace{-1.8cm}$^4$Australian Institute for Machine Learning, University of Adelaide \hspace{1.3cm} $^5$Valeo.ai \\
$^1${\tt\small \{firstname.lastname\}@sorbonne-universite.fr} \hspace{1cm}  $^3$ {\tt\small damien.teney@idiap.ch}
}

\maketitle

\footnotetext{\hspace{-0.2cm}\footnotemark[1]Equal contribution \hspace{0.1cm}\footnotemark[2]Work done before April 2021 and joining Tesla}
\ificcvfinal\thispagestyle{empty}\fi

\begin{abstract}
We introduce an evaluation methodology for visual question answering (VQA) to better diagnose cases of shortcut learning. These cases happen when a model exploits spurious statistical regularities to produce correct answers but does not actually deploy the desired behavior. There is a need to identify possible shortcuts in a dataset and assess their use before deploying a model in the real world. The research community in VQA has focused exclusively on question-based shortcuts, where a model might, for example, answer ``What is the color of the sky'' with ``blue'' by relying mostly on the question-conditional training prior and give little weight to visual evidence. We go a step further and consider multimodal shortcuts that involve both questions and images.
We first identify potential shortcuts in the popular VQA v2 training set by mining trivial predictive rules such as co-occurrences of words and visual elements. We then introduce VQA-CounterExamples (VQA-CE), an evaluation protocol based on our subset of CounterExamples i.e. image-question-answer triplets where our rules lead to incorrect answers. We use this new evaluation in a large-scale study of existing approaches for VQA. We demonstrate that even state-of-the-art models perform poorly and that existing techniques to reduce biases are largely ineffective in this context.
Our findings suggest that past work on question-based biases in VQA has only addressed one facet of a complex issue.
The code for our method is available at \url{https://github.com/cdancette/detect-shortcuts}
\end{abstract}

\vspace{-0.3cm}

\section{Introduction}

\begin{figure*}[h]
    \centering
    \includegraphics[width=0.9\linewidth]{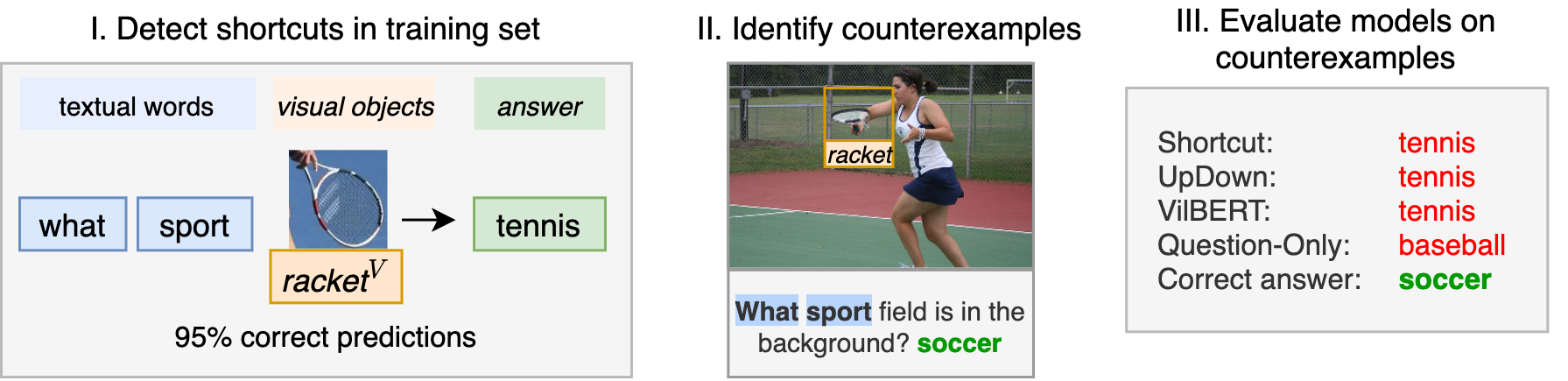}
    \caption{Overview of this work. We first mine simple predictive rules in the training data
    such as: \textual{what}~+~\textual{sport}~+~\visual{racket}~$\rightarrow$~\answer{tennis}.
    We then search for counterexamples in the validation set that identify some rules as undesirable statistical shortcuts. Finally, we use the counterexamples as a new challenging test set and evaluate existing VQA models like UpDown~\cite{anderson2018bottom} and VilBERT~\cite{lu2019vilbert}.}
    \label{fig:big_picture}
\end{figure*}

Visual Question Answering (VQA) is a popular task that aims at developing models able to answer free-form questions about the contents of given images.
The research community introduced several datasets~\cite{antol2015vqa, hudson2019gqa, johnson2017clevr, kafle2017tdiuc} to study various topics such as multimodal fusion~\cite{benyounes2017mutan} and visual reasoning~\cite{andreas2016neural, hu2017learning}.
The popular VQA v2 dataset~\cite{goyal2017vqa2} is the largest dataset of photographs of real scenes and human-provided questions. Because of strong selection biases and annotation artifacts, these datasets have served as a test-bed for the study of dataset biases and shortcut learning~\cite{geirhos2020shortcut} (we will use the term ``shortcut'' exclusively in the rest of the paper).
These spurious correlations correspond to superficial statistical patterns in the training data that allow predicting correct answers without deploying the desirable behavior.
Issues of shortcut learning have become an increasing concern for other tasks in vision and natural language processing~\cite{geirhos2020shortcut, d2020underspecification}.
In extreme cases, shortcuts in VQA may allow guessing the answer without even looking at the image~\cite{agrawal2018vqacp}.
Some shortcuts can be more subtle and involve both textual and visual elements.
For instance, training questions containing \textit{What sport} are strongly associated with the answer  \textit{tennis} when they co-occur with a racket in the image  (see Figure~\ref{fig:big_picture}).
However, some examples can be found in the validation set, such as \textit{What sport field is in the background ?}, that lead to a different answer (\textit{soccer}) despite a racquet being present in the image.
Because of such exceptions, a model that strongly relies on simple co-occurrences will fail on unusual questions and scenes.
Our work studies such multimodal patterns and their impact on VQA models.

The presence of dataset biases in VQA datasets is well known~\cite{agrawal2018vqacp, goyal2017vqa2, hudson2019gqa, kervadec2020gqaood}, but \textbf{existing evaluation protocols are limited to text-based shortcuts}.
Our work introduces VQA-\textit{CounterExamples} (VQA-CE for short) which is an evaluation protocol for multimodal shortcuts. It is easy to reproduce and can be used on any model trained on VQA v2, without requiring retraining.    
We first start with a method to discover superficial statistical patterns in a given VQA dataset that could be the cause of shortcut learning.
We discover a collection of co-occurrences of textual and visual elements that are strongly predictive of certain answers in the training data and often transfer to the validation set. For instance, we discover a rule that relies on the appearance of the words ``what'',``they'',``playing'' together with the object ``controller'' in the image to always predict the correct answer ``wii''.
We consider this rule to be a shortcut since it could fail on arbitrary images with other controllers, as it happens in the real world. Thus, our method can be used to reflect biases of the datasets that can potentially be learned by VQA models.
\vspace{-3mm}

We go one step further and identify counterexamples in the validation set where the shortcuts produce an incorrect answer. These counterexamples form a new challenging evaluation set for our VQA-CE evaluation protocol.
We found that the accuracy of existing VQA models is significantly degraded on this data.
More importantly, we found that most current approaches for reducing biases and shortcuts are ineffective in this context.
They often reduce the average accuracy over the full evaluation set without significant improvement on our set of counterexamples.
Finally, we identify shortcuts that VQA models may be exploiting. We find several shortcuts giving predictions highly correlated with existing models' predictions.
When they lead to incorrect answers on some examples from the validation set, VQA models also provide incorrect answers.
This tends to show that VQA models exploit these multimodal shortcuts.
In summary, the contributions of this paper are as follows.
 \vspace{-3mm}
\setlist{nolistsep,leftmargin=*}
\begin{enumerate}[noitemsep]
	\item We propose \textbf{a method to discover shortcuts} which rely on the appearance of words in the question and visual elements in the image to predict the correct answer. 
	By applying it to the widely-used VQA v2 training set, we found a high number of multimodal shortcuts that are predictive on the validation set.
	\item We introduce \textbf{the VQA-CE evaluation protocol} to assess the VQA models' reliance on these shortcuts.
	By running a large-scale evaluation of recent VQA approaches, we found that state-of-the-art models exploit these shortcuts and that bias-reduction methods are ineffective in this context.
\end{enumerate}

\section{Related Work}
\label{sec:related}

\begin{figure*}[h!]
    \centering
    \includegraphics[width=0.9\linewidth]{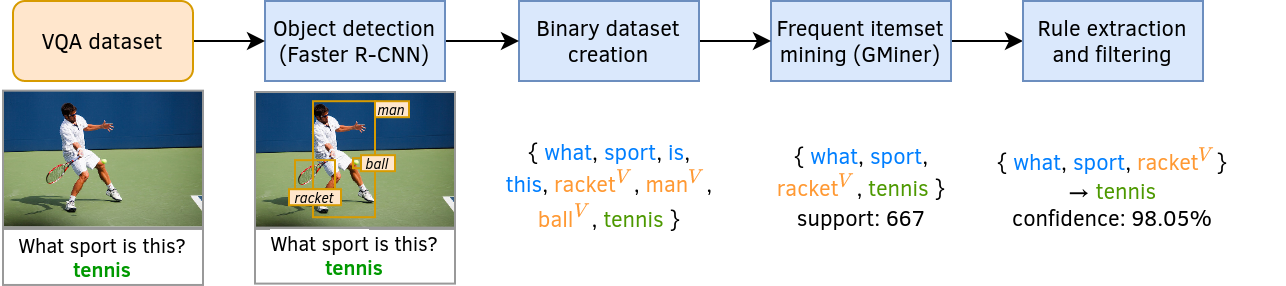}
    \caption{Pipeline of the proposed method to detect potential shortcuts in a VQA training set.
    We detect and label objects in images with a Faster R-CNN model.
    We then summarize each VQA example with binary indicators representing words in the question, answer, and labels of detected objects. Finally, a rule mining algorithm identifies frequent co-occurrences and extracts a set of simple predictive rules.}
    \label{fig:rule-mining}
\end{figure*}

We review existing approaches to discovering potential statistical shortcuts and assess their use by learned models.


\paragraph{Detecting cases of shortcut learning}
A first type of approaches consists in detecting the use of shortcuts by leveraging explainability methods~\cite{fong2017perturbation, ribeiro2016lime, stock2017imagenet, manjunatha2019explicit}. However, they require costly human interpretation, or additional annotations~\cite{das2017human} to assess whether a particular explanation reveals the use of a shortcut.
A second type consists in evaluating a model on artificial data or out-of-distribution data.
For instance, \cite{geirhos2018imagenet} artificially modify the texture of natural images to show that convolutional networks trained on ImageNet exploit features related to textures instead of shapes. 
Also, \cite{alcorn2019strike} and~\cite{barbu2019objectnet} evaluate vision models on out-of-distribution data to show that they cannot identify known objects when their poses changed significantly.
In this line of works, \cite{ilyas2019adversarial} focus on evaluating models on adversarial examples and show links with statistical regularities or ``non-robust features'' that models exploit.
A third type of approaches use specific models with a known type of biases to assess the amount of biases of this type directly in the dataset.
For instance, in computer vision, BagNet~\cite{brendel2018approximating} obtained high accuracy on ImageNet by using occurrences of small local image features, without using the global spatial context. This suggests that state-of-the-art ImageNet models are biased towards local image features.
Similarly, our approach leverages specific shallow models that are constructed to only exploit biases of a certain type.

This kind of approaches have been used in VQA. Previous works~\cite{agrawal2018vqacp, antol2015vqa, goyal2017vqa2} used question-only and image-only models to quantify the amount of unimodal shortcuts in a dataset. Instead, our approach is not only able to quantify the amount of shortcuts but also identify these shortcuts. More importantly, our method can identify multimodal shortcuts that combine elements of the question and the image.
The closest approach to ours~\cite{manjunatha2019explicit} uses the Apriori algorithm to extract predictive rules that combine the appearance of words and visual contents. However, these rules are specific to the attention maps and predictions of the VQA model from~\cite{kazemi2017show}. More problematically, they are extracted on the validation set and are mainly used for qualitative purposes. 
Our approach also relies on the Apriori algorithm but extracts rules directly on the training set, independently of any model, and the predictive capacity of the rules is evaluated on the validation set.

\vspace{-2mm}
\paragraph{Evaluating VQA models' reliance on shortcuts}
Once a class of shortcuts has been identified, a first way to evaluate model's robustness is to build external out-of-distribution evaluation datasets on which using these shortcuts leads to a wrong prediction.
In Visual Question Answering, the VQA-Rephrasing~\cite{meet2019cycle} dataset contains multiple rephrased but semantically-identical questions. The goal is to test model's sensitivity to small linguistic variations and will penalize usage of a certain class of question-related shortcuts. Similar datasets exist for natural language processing~\cite{jia2017adversarialsquad,mccoy2019-right-wrong-hans}.

Another type of evaluation methods artificially injects certain kind of shortcuts in the training set and evaluate models on examples that do not possess these shortcuts.
The widely used \textbf{VQA-CP}~\cite{agrawal2018vqacp} evaluation procedure consists in resplitting the original VQA datasets so that the distribution of answers per question type (``how many'', ``what color is'', etc.) is different between the training and evaluation set. Models that rely on those artificial shortcuts are therefore penalized. VQA-CP was used to develop methods that aim at avoiding learning shortcuts from the question type on this modified training set~\cite{cadene2019rubi, clark2019ensemblebias, gat2020rmfe, ramakrishnan2018overcoming, gokhale2020mutant, chen2020counterfactual, gat2020rmfe, shrestha2020negativecase, teney2020unshuffling, teney2019actively, teney2020value}.
Similar approaches for VQA exists~\cite{clark2019ensemblebias}.
The downside of these approaches is that they focus on artificially introduced shortcuts and only target text-related biases and shortcuts. More importantly, models that have been trained on original datasets, i.e. VQA v2, need to be retrained on their modified versions, i.e. VQA-CP v2. Other concerns have been raised in~\cite{teney2020value}.
On the contrary, our proposed evaluation method does not require additional data collection or data generation, focuses on multimodal shortcuts, and does not require retraining. We follow guidelines from~\cite{d2020underspecification,teney2020value} for a better evaluation of the use of shortcuts.

Finally, the GQA-OOD \cite{kervadec2020gqaood} dataset extracts from the GQA\cite{hudson2019gqa} validation and testing set example with rare answers, conditioned on the type of question. Thus, it targets question-related shortcuts. It enables the evaluation of models without retraining on a separate training set.



\section{Detecting multimodal shortcuts for VQA}
\label{sec:discover}

\begin{figure*}[h!]
    \centering
    \includegraphics[width=0.95\linewidth]{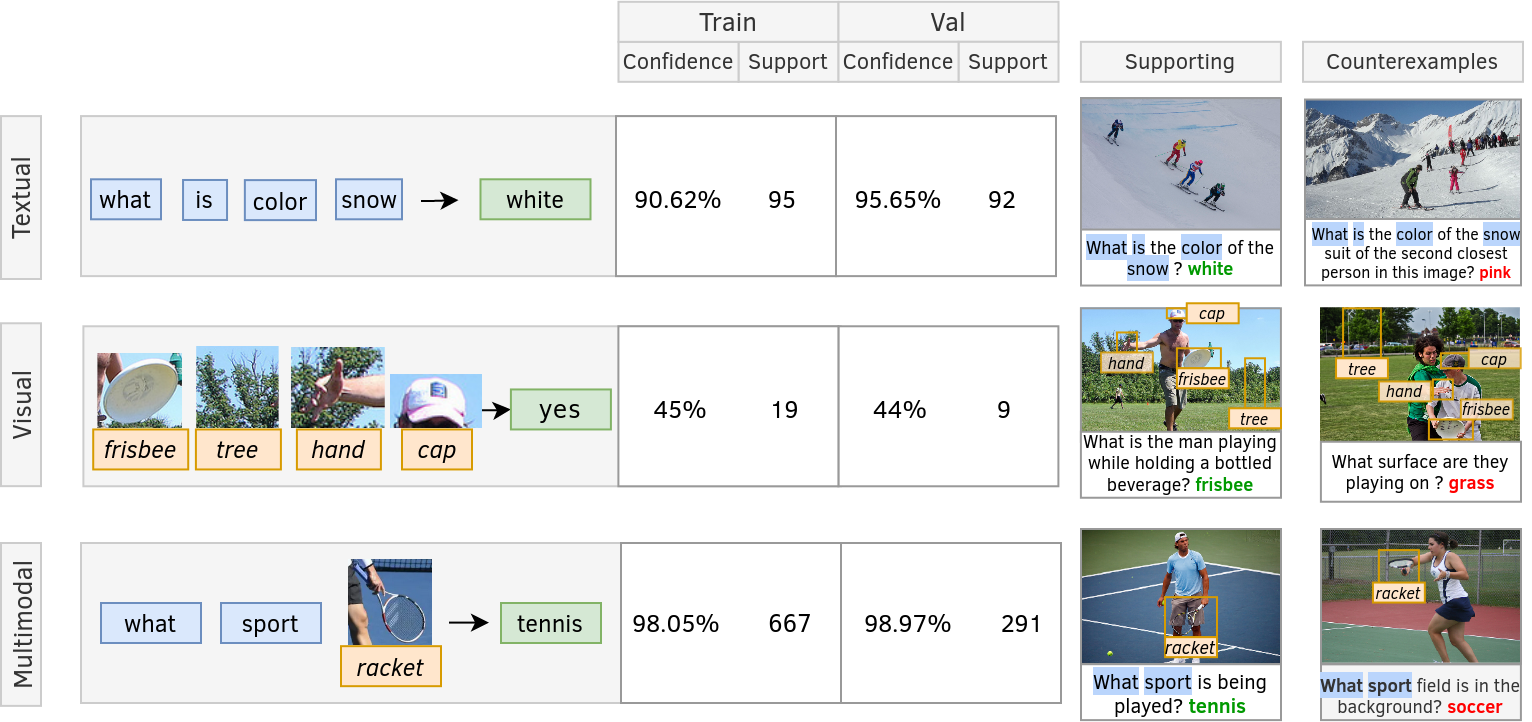}
    \caption{Examples of shortcuts found in the VQA v2 dataset. The confidence is the accuracy obtained by applying the shortcut on all examples matching by its \textit{antecedent}. The support is the number of matching examples. More supporting examples and counterexamples are shown in the supplementary material.}
    \vspace{-2mm}
    \label{fig:rules-vqa2}
\end{figure*}

\subsection{Our shortcut detection method}
We introduce our method to detect shortcuts relying on textual and visual input. 
Our approach consists in building a dataset of input-output variables and applying a rule mining algorithm. 
The code for our method is available online \footnote{ \url{https://github.com/cdancette/detect-shortcuts}}.
In \textbf{}Visual Question Answering (VQA), we consider a training set $\mathcal{D}_{train}$ made of $n$ triplets $(\vi,\qi,\ai)_{i \in [1,n]}$ with $\vi \in \mathcal{V}$ an image, $\qi \in \mathcal{Q}$ a question in natural language and $\ai \in \mathcal{A}$ an answer.
VQA is usually casted as a problem of learning a multimodal function $f: \mathcal{V} \times \mathcal{Q} \rightarrow \mathcal{A}$ that produces accurate predictions on $\mathcal{D}_{test}$ of unseen triplets.

\vspace{-3mm}
\paragraph{Mining predictive rules on a training set}
Our goal is to detect shortcuts that $f$ might use to provide an answer without deploying the desired behavior.
To this end, we limit ourselves to a class of shortcuts that we think is often leveraged by $f$. We display in Figure~\ref{fig:rule-mining} our rule mining process.
These shortcuts are short predictive association rules $A \rightarrow C$ that associate an \textbf{antecedent} $A$ to a \textbf{consequent} $C$. Our antecedents are composed of words of the question and salient objects in the image (or image patch), while our consequents are just answers. For instance, the rule \{\textual{what}, \textual{color}, \textual{plant}\} $\rightarrow$ \{\answer{green}\} provides the answer ``green'' when the question contains the words ``what'', ``color'' and ``plant''.
These shallow rules are by construction shortcuts. They are predictive on the validation set but do not reflect the complex behavior that needs to be learned to solve the VQA task.
For instance, they do not rely on the order of words, neither the position and relationships of visual contents in the image. They lack the context that is required to properly answer the question.
Moreover, even rules that seem correct often have a few counterexamples in the dataset, i.e. examples that are matched by the antecedent but the consequent provides the wrong answer. We later use these counterexamples in our evaluation procedure.

\vspace{-4mm}
\paragraph{Binary dataset creation}
To detect these rules, we first encode all question-image-answer triplets of $\mathcal{D}_{train}$ as binary vectors.
Each dimension accounts for the presence or absence of \textbf{(a)} a \textual{word} in the question, \textbf{(b)} an \visual{object} in the image, represented by its textual detection label from Faster R-CNN, \textbf{(c)} an \answer{answer}. The number of dimensions of each binary vector is the sum of the size of the dictionary of words (e.g. \textapprox 13,000 words in VQA v2), the number of detection labels of distinct objects in all images (e.g. 1,600 object labels), and the number of possible answers in the training set (e.g. 3,000 answers). We report results with ground-truth instead of detected labels in the supplementary materials.

\vspace{-4mm}
\paragraph{Frequent itemset mining}
On our binary dataset, we apply the GMiner algorithm~\cite{chon2018gminer} to efficiently find frequent \textit{itemsets}. An itemset is a set of tokens $\mathcal{I} = \{i_1, .., i_n\}$ that appear very frequently together in the dataset. The \textbf{support} of the itemset is its number of occurrences.
For example, the itemset \{\textual{what}, \textual{color}, \textual{plant}, \answer{green}\} might be very common in the dataset and have a high support.
GMiner takes one parameter, the minimum support. We include an additional parameter, which is the maximum length for an itemset. We detail how we select parameters at the end of this section. 

\vspace{-4mm}
\paragraph{Rules extraction and filtering}
The next step is to extract rules from the frequent itemsets.
First, we filter out the itemsets that do not contain an answer token, as they cannot be converted to rules.
For the others that do contain an answer $a$, we remove it from the itemset to create the antecedent $\mathcal{X}$ ( $\mathcal{X} = \mathcal{I} \setminus a$).
The rule is then $\mathcal{X} \Rightarrow a $.
The \textbf{support} $s$ of the rule is the number of occurrences of $\mathcal{X}$ in the dataset.
The \textbf{confidence} $c$ of the rule is the frequency of correct answers among examples that have $\mathcal{X}$.
\\
We then proceed to filter rules.
We apply the following three steps: 
\textbf{(a)} we remove the rules with a confidence on the training set lower than 30\% (remove when $c < 0.3$)~~
\textbf{(b)} if some rules have the same antecedent but different answers, then we keep the rule with the highest confidence and remove the others. For instance, given the rules \{\textual{is}, \textual{there}\} $\Rightarrow$ \answer{yes} and  \{\textual{is}, \textual{there}\} $\Rightarrow$ \answer{no} with a respective confidence of 70\% and 30\%, we only keep the first one with the answer $\answer{yes}$.~~
\textbf{(c)} if a rules $r_1$'s antecedent is a superset of another rule $r_2$'s antecedent, if both have the same answer, and $r_1$ has a lower confidence than $r_2$, then we remove $r_1$.
For instance,  given the rules \{\textual{is}, \textual{there}\} $\Rightarrow$ \answer{yes} and  \{\textual{is}, \textual{there}, \textual{cat}\} $\Rightarrow$ \answer{yes} with a respective confidence of 70\% and 60\%, we only keep the first one without the word $\textual{cat}$. We consider the remaining rules as shortcuts. Note that rules with a confidence of 100\% could be considered \textit{correct} and not shortcuts, but these rules will not influence our evaluation protocol, detailed in Section~\ref{sec:evaluation}.

\vspace{-2mm}
\subsection{Analysis of shortcuts on natural data}
We analyze the shortcuts that our approach can detect on the widely used VQA v2 dataset~\cite{goyal2017vqa2} made of ~1.1M image-question-answer examples and based on ~200K images from the MS-COCO dataset~\cite{lin2014microsoftcoco}.
We extract shortcuts with different combinations of minimum support and confidence. Each time, we aggregate them into a classifier that we evaluate on the validation set. We detail how to build this kind of classifier in Section~\ref{sec:model-evaluation}. We select the support and confidence leading to the best overall accuracy.
It corresponds to a minimum support of $2.1\cdot 10^{-5}$ (about $\sim$8 examples in training set), and a minimum confidence of 0.3.
Once these shortcuts have been detected, we assess their number and type(purely textual, purely visual, or multimodal). We also verify that they can be used to find counterexamples that cannot be accurately answered using shortcuts.
Finally, we evaluate their confidence on the validation set.
In the next section, we leverage these counterexamples with our VQA-CE evaluation protocol to assess model's reliance on shortcuts.

\vspace{-3mm}
\paragraph{Words-only and objects-only shortcuts}
First, we show that our approach is able to detect shortcuts that are purely textual or visual. In the first row of Figure~\ref{fig:rules-vqa2}, we display a shortcut detected on VQA v2 that only accounts for the appearance of words in the question. It predicts the answer ``white'' when the words ``what'', ``color'', ``is'', ``snow'' appear at any position in the question. In the training set, these words appear in 95 examples and 90.62\% of them have the ``white'' answer. This shortcut is highly predictive on the validation set and gets 95.65\% of correct answers over 92 examples. We also display an example on which exploiting the shortcut leads to the correct answer, and a counterexample on which the shortcut fails because the question was about ``the color of the snow suit'' which is ``pink''.
In the second row, we show a shortcut that only accounts for the appearance of visual objects. It predicts ``yes'' when a ``frisbee'', a ``tree'', a ``hand'' and a ``cap'' appear in the image. However, this kind of shortcuts is usually less predictive since they cannot exploit the question-type information which is highly correlated with certain answers, i.e. ``what color'' is usually answered by a color.

\vspace{-4mm}
\paragraph{Multimodal shortcuts}
Then, we show that our approach is able to detect multimodal shortcuts.
They account for the appearance of both \textual{words} \textbf{and} visual \visual{objects}.
In the third row of Figure~\ref{fig:rules-vqa2}, we display a multimodal shortcut that predicts ``tennis'' when the words \textual{what}, \textual{sport} and a \visual{racket} appear. It is a common shortcut with a confidence of 98.05\% based on a support of 667 examples in the training set. It is also highly predictive on the validation set with 98.97\% confidence and 291 support. 
At first sight, it is counter-intuitive that this simple rule is a shortcut but answering complex questions is not about detecting frequent words and objects in images that correlate with an answer.
In fact, this shortcut is associated to counterexamples where it fails to answer accurately. Here, the sport that can be played in the background is not tennis but soccer.

\begin{figure}[h!]
    \centering
    \includegraphics[width=0.9\linewidth]{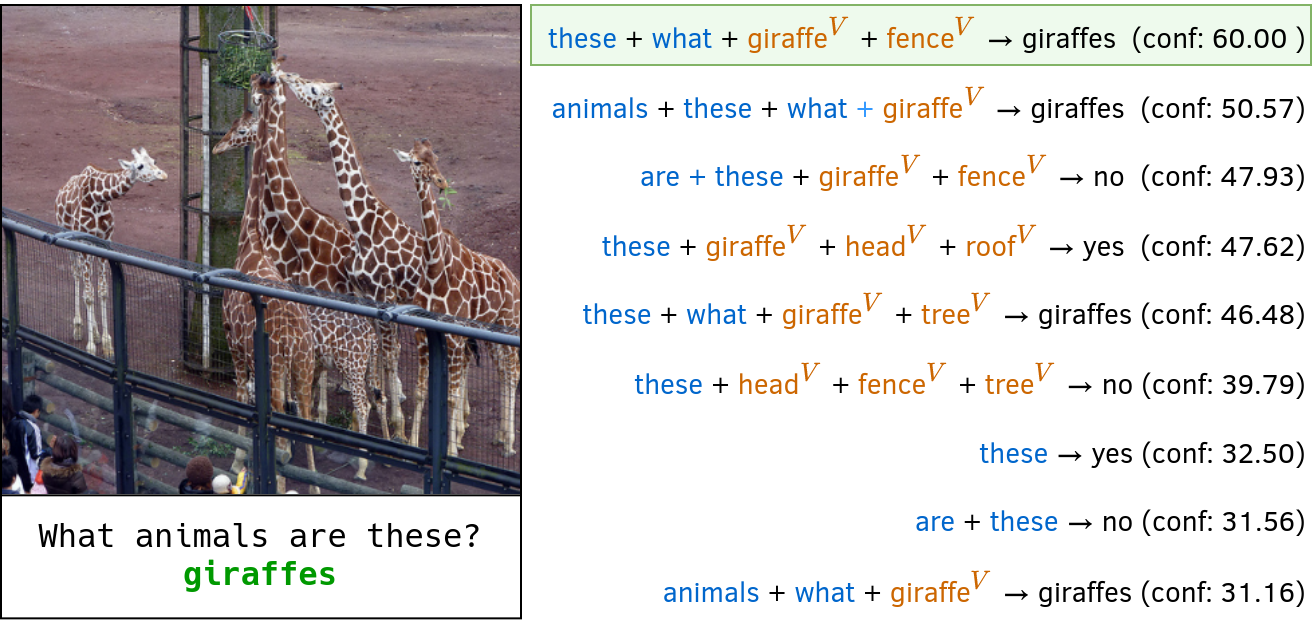}
    \caption{Multiple shortcuts can often be exploited to find the correct answer in any given example.
    The confidence is the percentage of accurate answers among examples that are matched by the shortcut \textit{antecedent}.
    The shortcut of highest confidence (in green) is multimodal for \textapprox 92\% of examples.\vspace{-3mm}}
    \label{fig:rule-selection}
\end{figure}

\vspace{-4mm}
\paragraph{Number of shortcuts and statistics per type}
Here we show that our approach can be used to detect a high number of multimodal shortcuts.
Overall, it detects \textapprox1.12M shortcuts on the VQA v2 training set.
As illustrated in Figure~\ref{fig:rule-selection}, since there are \textapprox413K examples, it is often the case that several shortcuts can be applied to the same example. This is the main reason behind the high number of shortcuts 
For instance, the antecedent \{\textual{animals}, \textual{what}, \visual{giraffe}\} overlaps with \{\textual{animals}, \textual{these}, \textual{what}, \visual{giraffe}\}.
Among all the shortcuts that our method can detect, only \textapprox50k are textual, \textapprox77k are visual and \textapprox1M are multimodal. In other words, \textapprox90\% are multimodal. In addition to being more numerous, they are also more predictive.
For instance, the most confident shortcut that matches an example, highlighted in green in Figure~\ref{fig:rule-selection}, is multimodal 91.80\% of the time.
Finally, \textapprox3K examples are not matched by any shortcut antecedents. They have unusual question words or visual content. 
We later do not take them into account in our VQA-CE evaluation protocol. We display some representative examples in the supplementary materials.

\vspace{-4mm}
\paragraph{Confidence distribution on training and unseen data}
In the supplementary materials, we display the confidence distribution of these shortcuts. 
We observe that our shortcuts are predictive on unseen data that follows the training set distribution. The number of shortcuts that reach a confidence between 0.9 and 1.0 is as high on the validation set as it is than on the training set.
This means that shortcuts detected on the VQA v2 training set transfer to the validation set.
Additionally, most shortcuts obtain a confidence lower than 1.0, which allows finding examples that contradict them by leading to the wrong answers.
These counterexamples are the core of our approach to assess the VQA model's reliance on shortcuts which is described next.



\vspace{-1mm}
\section{Assessing models' reliance on shortcuts}
\label{sec:evaluation}

The classic evaluation protocol in VQA consists in calculating the average accuracy over all the examples.
Instead, we introduce the VQA-\textit{CounterExamples} evaluation protocol (VQA-CE) that additionally calculates the average accuracy over a specific subset of the validation set. This subset is made of counterexamples that cannot be answered by exploiting shortcuts. Models that do exploit shortcuts are expected to get a lower accuracy.
It is how we assess the use of shortcuts.
Importantly, our protocol does not require retraining as it was the case with the previous VQA-CP~\cite{agrawal2018vqacp} protocol.
We first detail the subsets creation procedure at the core of our VQA-CE protocol.
Then we run extensive experiments to assess the use of shortcuts on many VQA models and bias-reduction methods.
Finally, we identify shortcuts that are often exploited by VQA models.

\vspace{-1mm}
\subsection{Our VQA-CE evaluation protocol}
\paragraph{Subsets creation using shortcuts}
By leveraging the shortcuts that we have detected before, we build the \textbf{Counterexamples} subset of the VQA v2 validation set.
This subset is made of 63,298 examples on which all shortcuts provide the incorrect answer.
As a consequence, VQA models that exploit these shortcuts to predict will not be able to get accurate answers on this kind of examples. They will be penalized and obtain a lower accuracy on this subset.
On the contrary, we build the non-overlapping \textbf{Easy} subset.
It is made of 147,681 examples on which at least one shortcut provides the correct answer.
On this subset, VQA models that exploit shortcuts can reach high accuracy.
Finally, 3,375 examples are not matched by any shortcut's antecedent. Since these examples do not belong to any of our two subsets, we do not consider them in our analysis.
We show in supplementary materials that they have unusual questions and images.

\begin{figure}[h!]
    \centering
    \includegraphics[width=1.0\linewidth]{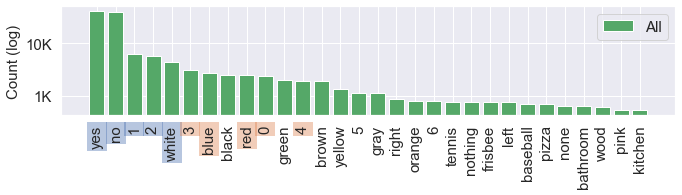}
    \includegraphics[width=1.0\linewidth]{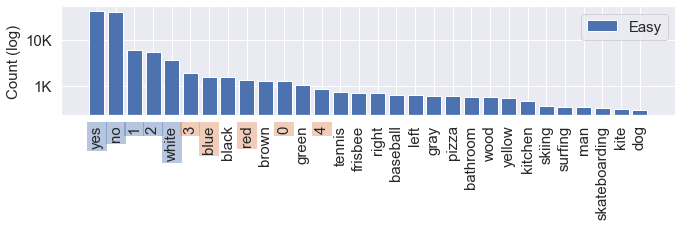}
    \includegraphics[width=1.0\linewidth]{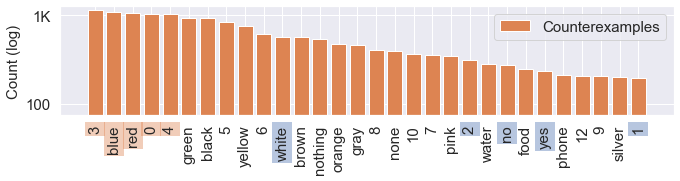}
    \caption{Number of examples per answer (30 most frequent ones) in the complete validation set, our Counterexamples subset, and our Easy subset. Answers highlighted in blue and orange are the top 5 answers for the Easy and Counterexamples subsets respectively. \vspace{-2mm}}
    \label{fig:distrib-answers}
\end{figure}

\vspace{-3.5mm}
\paragraph{Distribution of examples}
Here, we show how the split between our two subsets Counterexamples and Easy affects the distribution of examples.
In Figure~\ref{fig:distrib-answers}, we show that the original distribution of answers is similar to the Easy distribution but dissimilar to the Counterexamples distribution.
Highlighted in blue, we display the five most common answers from the Easy distribution. They can be found at the same positions in the original distribution, the two major answers being ``yes'' and ``no''.
It is not the case in the Counterexamples subset where these answers appear less frequently. Nonetheless, they are still in the top 30 answers which shows that our subsets creation is not a trivial splitting between frequent and rare answers.
Similarly, the five most common answers from the Counterexamples subset, highlighted in orange, can be found in the Easy and All subset.
We report similar observations for the questions and answer-type distributions in the supplementary materials.

\vspace{2mm}
\subsection{Main results}
\label{sec:model-evaluation}

\begin{table*}[h!]
    \small
    \centering
    \begin{tabular}{c p{4cm} P{2cm} P{4cm} P{2cm}}
    \toprule
    & Approaches & Overall & \textbf{Counterexamples (ours)} & \textbf{Easy (ours)}\\
    & \multicolumn{1}{r}{\scriptsize \textit{Number of examples}} &  {\scriptsize \textit{214,354}} & {\scriptsize \textit{63,298}} & {\scriptsize \textit{147,681}} \\ 
    \midrule
    {\multirow{3}{*}{\rotatebox[origin=c]{90}{\footnotesize Baselines}}}
    & Shortcuts & 42.26 \white{(+0.00)} & 0.00 \white{(+0.00)} & 61.13 \white{(+0.00)}\\
    & Image-Only & 23.70 \white{(+0.00)} & 1.58 \white{(+0.00)} & 33.58 \white{(+0.00)}  \\
    & Question-Only & 44.12 \white{(+0.00)} & 11.59 \white{(+0.00)} & 58.61 \white{(+0.00)}\\
    \midrule
    \multirow{4}{*}{\rotatebox[origin=c]{90}{\footnotesize VQA models}} 
    & SAN~\cite{yang2016stacked} -- \textit{grid features} & 55.61 \white{(+0.00)} & 26.64 \white{(+0.00)} & 68.45 \white{(+0.00)} \\
    &UpDown~\cite{anderson2018bottom} & 63.52  \positive{(+0.00)}  & 33.91  \positive{(+0.00)}  & 76.69  \positive{(+0.00)}  \\
    & BLOCK~\cite{benyounes2019block} & 63.89 \white{(+0.00)} & 32.91 \white{(+0.00)} & 77.65 \white{(+0.00)} \\
    & VilBERT~\cite{lu2019vilbert}  -- \textit{pretrained}$^\dagger$   & 67.77 \white{(+0.00)} & 39.24 \white{(+0.00)} & 80.50 \white{(+0.00)} \\
    \midrule
    \multirow{8}{*}{\rotatebox[origin=c]{90}{{\footnotesize Bias-reduction methods}}} 
    & \multicolumn{4}{c}{\textit{UpDown~\cite{anderson2018bottom} is used as a base architecture for bias-reduction methods}} \\
    & RUBi~\cite{cadene2019rubi} &  61.88 \negative{(-1,64)} & 32.25 \negative{(-1,66)} & 75.03 \negative{(-1.66)} \\
    & LMH + RMFE~\cite{gat2020rmfe} & 60.96 \negative{(-2.56)} & 33.14 \negative{(-0.77)} & 73.32 \negative{(-3.37)} \\
    & ESR~\cite{shrestha2020negativecase} & 62.96 \negative{(-0.56)} & 33.26 \negative{(-0.65)} & 76.18  \negative{(-0.51)}\\
    & LMH~\cite{clark2019ensemblebias} & 61.15 \negative{(-2.37)} & 34.26 \positive{(+0.35)} & 73.12  \negative{(-3.57)}\\
    & LfF~\cite{nam2020learningfromfailure} & 63.57 \positive{(+0.05)} & 34.27 \positive{(+0.36)} & 76.60 \negative{(-0.09)} \\
    & LMH+CSS~\cite{chen2020counterfactual} & 53.55 \negative{(-9.97)} & 34.36 \positive{(+0.45)} & 62.08 \negative{(-14.61)} \\
    & RandImg~\cite{teney2020value} & 63.34 \negative{(-0.18)} & 34.41 \positive{(+0.50)} & 76.21 \negative{(-0.48)}\\
    \bottomrule \\
    \end{tabular}
    \hspace{5mm}
    \begin{tabular}{c}
    \toprule
    VQA-CP v2 \cite{agrawal2018vqacp} \\
    \\
    \midrule
    22.64 \\  
    19.31 \\ 
    15.95 \\ 
    \midrule
    24.96 \\  
    39.74 \\ 
    38.69 \\ 
    | \\
    \midrule
    \\
    44.23 \\ 
    54.55 \\ 
    48.50 \\  
    52.05 \\ 
    39.49  \\ 
    58.95 \\ 
    55.37 \\  
    \bottomrule \\
    \end{tabular}
    \vspace{-1mm}
    \caption{Results of our VQA-CE evaluation protocol. We report accuracies on VQA v2 full validation set and on our two subsets: \textbf{Counterexamples} and \textbf{Easy} examples. We re-implemented all models and bias-reduction methods.
    $^\dagger$VilBERT is pretrained on Conceptual Caption and fine-tuned on VQA v2 training set. Scores in {\positive{(green)}} and {\negative{(red)}} are relative to UpDown~\cite{anderson2018bottom}.
    We also report accuracies on VQA-CP v2~\cite{agrawal2018vqacp} which focus on question biases, and comes with a different training set and testing set. VilBERT was not evaluated for VQA-CP as it was pretrained on balanced datasets. \vspace{-1mm}}
    \vspace{-2mm}
    \label{tab:vqa2-results-counterfactual}
\end{table*}

In Table~\ref{tab:vqa2-results-counterfactual}, we report results of some baselines, common VQA models, and latest bias-reduction methods following our VQA-CE evaluation protocol. Models that exploit shortcuts are expected to get a lower accuracy on the Counterexamples compared to their overall accuracy. All models have been trained on the VQA v2 training set, and evaluated on the VQA v2 validation set. We detail them and discuss our findings in the next paragraphs. We report additional results on VQA v1 and v2 in the supplementary materials.

\vspace{-3mm}
\paragraph{Baselines}
The Question-Only and Image-Only baselines are deep models that only use one modality.
They are often used to assess the amount of unimodal shortcuts that a deep model can capture.
We report extreme drops in accuracy on our Counterexamples subset compared to the overall accuracy, with a loss of 32.53 points and 22.12 points respectively. This shows that most of the questions that are easily answerable by only using the question, or the image, are filtered out of our Counterexamples subset.

\vspace{-3mm}
\paragraph{Aggregating shortcuts to create a classifier}
In order to evaluate our shortcuts as a whole, we aggregate them to build a VQA classifier.
As shown in the preceding section, each training example is associated with shortcuts that can be used to find the correct answer.
Among these correct shortcuts, we select the highest-confidence one for each example. This leaves us with 115,718 unique shortcuts.
In order to predict an answer for an unseen example, we take the most predicted answer for all its matching shortcuts weighted by the confidence of the shortcuts.
For the examples that are not matched by any shortcut, we output ``yes'', the most common answer.
Our shortcut-based classifier reaches an overall accuracy of 42.26\%, close to the 44.12\% of the deep question-only baseline. Interestingly, both use a different class of shortcuts. Ours is mostly based on shallow multimodal shortcuts, not just shortcuts from the question.
Since we use the same shortcuts to create our subsets, the shortcut-based classifier reaches a score of 0\% on the Counterexamples.
On VQA-CP testing set, our classifier reaches 22.44\% accuracy. It highlights the difference with our counterexamples subset: VQA-CP does penalize some shortcuts, but there are still some that can be exploited.

\begin{table*}[t]
    \small
    \centering
    \begin{tabular}{rllcccc}
    \toprule
    & \multicolumn{1}{c}{Train} & \multicolumn{1}{c}{Val} &  \multicolumn{3}{c}{Correlations (Val)} \\
    \cmidrule(l){4-6} 
    \multicolumn{1}{c}{Rule (antecedent $\rightarrow$ consequent)} & \multicolumn{1}{c}{Conf. (Sup.)} & \multicolumn{1}{c}{Conf. (Sup.)} & UpDown & VilBERT & Question-Only \\
    \midrule
    \textual{doing} + \visual{man} + \visual{surfboard} + \visual{hand} $\rightarrow$ \answer{surfing} & 86.6 (115) & 87.3 (55) & 100.0 & 100.0 & 23.6 \\
    \textual{sport} + \textual{this} + \textual{what} + \visual{skateboard} $\rightarrow$ \answer{skateboarding} & 98.2 (53) & 87.1 (31) & 100.0 & 100.0 & 0.0 \\
    \textual{holding} + \textual{this} + \textual{what} + \visual{racket} $\rightarrow$ \answer{tennis racket} & 75.0 (26) & 33.3 (3) & 100.0 & 100.0 & 33.3 \\
    \textual{played} + \visual{shorts} + \visual{racket} + \visual{leg} $\rightarrow$ \answer{tennis} & 100.0 (29) & 80.0 (5) & 100.0 & 100.0 & 40.0 \\
    \textual{playing} + \textual{they} + \textual{what} + \visual{controller} $\rightarrow$ \answer{wii} & 100.0 (30) & 88.9 (9) & 100.0 & 100.0 & 66.7 \\
    \textual{picture} + \textual{where} + \visual{beach} + \visual{people} $\rightarrow$ \answer{beach} & 100.0 (21) & 90.0 (10) & 100.0 & 100.0 & 90.0 \\
    \textual{taken} + \textual{where} + \visual{toilet} $\rightarrow$ \answer{bathroom} & 85.2 (22) & 80.0 (5) & 100.0 & 100.0 & 20.0 \\
    \textual{eating} + \textual{what} + \visual{pizza} + \visual{arm} $\rightarrow$ \answer{pizza} & 81.5 (21) & 66.7 (6) & 100.0 & 100.0 & 66.7 \\
    \textual{carrying} + \textual{is} + \textual{what} + \visual{kite} $\rightarrow$ \answer{kite} & 66.7 (21) & 60.0 (5) & 100.0 & 100.0 & 0.0 \\
    \textual{gender} + \textual{of} + \textual{what} + \visual{head} $\rightarrow$ \answer{male} & 64.1 (24) & 66.7 (6) & 100.0 & 100.0 & 66.7 \\
    \textual{position} + \visual{helmet} + \visual{bat} + \visual{dirt} $\rightarrow$ \answer{batter} & 61.8 (20) & 71.4 (7) & 100.0 & 100.0 & 0.0 \\
    \bottomrule
    \end{tabular}
    \vspace{0.23cm}
    \caption{Instances of shortcuts that are highly correlated with VQA models' predictions. We display their antecedent made of \textual{words} from the question and \visual{objects} from the image, and their \answer{answer}. Their support, i.e. number of examples matched by the antecedent, and confidence, i.e. percentage of correct answers among them, have been calculated on the VQA v2 training and validation sets. We report the correlation coefficients of their predictions with those of three VQA models: UpDown~\cite{anderson2018bottom} that uses an object detector, VilBERT~\cite{lu2019vilbert} that has been pretrained on a large dataset, and Q-only~\cite{goyal2017vqa2} that only uses the question. We show some counterexamples in the supplementary material. \vspace{-4mm}}
    \label{tab:example-shortcuts}
\end{table*}

\vspace{-2mm}
\paragraph{VQA models learn shortcuts}
We compare different types of VQA models: SAN~\cite{yang2016stacked} represents the image as a grid of smaller patches and uses a stacked attention mechanism over these patches,
instead UpDown~\cite{anderson2018bottom} represents the image as a set of objects detected with Faster-RCNN and uses a simpler attention mechanism over them,
BLOCK~\cite{benyounes2019block} also relies on the object representations but uses a more complex attention mechanism based on a bilinear fusion,
VilBERT~\cite{lu2019vilbert} also relies on the object representations but uses a transformer-based model that has been pretrained on the Conceptual Caption dataset~\cite{sharma2018conceptual}.
First, they suffer from a loss of \textapprox29 accuracy points on the counterexamples compared to their overall accuracy.
This suggests that, despite their differences in modeling, they all exploit shortcuts.
Note that comparable losses are reported on VQA-CP~v2~\cite{agrawal2018vqacp} which especially focuses on shortcuts based on question-types.
Second, our evaluation protocol can be used to compare two models that get similar overall accuracies: UpDown and BLOCK which gets +0.37 points over UpDown.
We can explain that this gain is due to a superior accuracy on the Easy subset with +0.96 and report a loss of -1.00 points on the Counterexamples. These results suggest that the bilinear fusion of BLOCK better captures shortcuts.
On the contrary, VilBERT gets a better accuracy on our both subsets. It might be explained by the advantages of pretraining on external data.

\vspace{-2mm}
\paragraph{Bias-reduction methods do not work well on natural multimodal shortcuts}
Our evaluation protocol can also be used to assess the efficiency of common bias-reduction methods. We use publicly available codebases when available, or our own implementation.
All methods have been developed on the VQA-CP v2 dataset. It introduces new training and evaluation splits of VQA v2 that follow different distributions conditioned on the question-type.
All the studied methods have been applied to UpDown and reached gains ranging from +5 to +20 accuracy points on the VQA-CP testing set. 
We evaluate them in the more realistic context of the original VQA v2 dataset.
We show that their effect on our Counterexamples subset is very small. 
More specifically, some methods such as RUBi~\cite{cadene2019rubi}, LMH+RMFE~\cite{gat2020rmfe}, and ESR~\cite{shrestha2020negativecase} have a negative effect on all subsets.
Some methods such as LMH~\cite{clark2019ensemblebias} and LMH+CSS~\cite{chen2020counterfactual} slightly improve the accuracy on counterexamples but strongly decrease the accuracy on the Easy subset, and consequently decrease the overall accuracy.
As reported in~\cite{teney2020value}, most of those methods rely on knowledge about the VQA-CP testing distribution (inversion of the answer distribution conditioned on the question), which no longer applies in our VQA v2 evaluation setting.
Finally, we found two methods, LfF~\cite{nam2020learningfromfailure} and RandImg~\cite{teney2020value} that slightly improve the accuracy on the Counterexamples subset with gains of +0.36 and +0.50, while having a very small impact on the overall accuracy, even reaching small gains in the best case of LfF. It should be noted that LfF is more general than others since it was not designed for the VQA-CP context.
Overall, all effects are much smaller compared to their effectiveness on VQA-CP.
This suggests that those bias-reduction methods might exploit the distribution shift between VQA-CP training and evaluation splits. They are efficient in this setting but do not work as well to reduce \textit{naturally-occurring} shortcuts in VQA.

\subsection{Identifying most exploited shortcuts}

We introduce a method to identify shortcuts that may be exploited by a given model.
On the validation set, we calculate for each shortcut a correlation coefficient between its answer and the predictions of the studied model.
Importantly, a 100\% correlation coefficient indicates that the model may exploit the shortcut: 
both always provide the same answers, even on counterexamples on which using the shortcuts leads to the wrong answer.

In Table~\ref{tab:example-shortcuts}, we report shortcuts that obtain the highest correlation coefficient with UpDown~\cite{anderson2018bottom} and VilBERT~\cite{lu2019vilbert}.
Overall, these shortcuts have a high confidence and support, which means that they are common in the dataset and predictive.
Most importantly, they are multimodal.
As a consequence, these shortcuts obtain low correlations with Question-Only~\cite{goyal2017vqa2}.
On the contrary, they obtain a 100\% correlation coefficient with VilBERT and UpDown.
For instance, the second shortcut provides the answer \answer{skateboarding} for the appearance of \textual{sport}, \textual{this}, \textual{what} in the question and a \visual{skateboard} in the image.
It is a common shortcut with a support of 31 examples in the validation set. It gets a correlation of 0\% because Question-Only mostly answer baseball for these examples.
Its confidence of 87.1\% indicates that 4 counterexamples can be found where the shortcut provides the wrong answer. To be correctly answered, they require more than a simple prediction based on the appearance of words and salient visual contents.
These results once again confirm that VQA models tend to exploit multimodal shortcuts. It shows the importance of taking them into account in an evaluation protocol for VQA.


\vspace{-2mm}
\section{Conclusion}
\vspace{-2mm}
We introduced a method that discovers multimodal shortcuts in VQA datasets. It gives novel insights on the nature of shortcuts in VQA: they are not only related to the question but are also multimodal. 
We introduced an evaluation protocol to assess whether a given models exploits multimodal shortcuts.
We found that most state-of-the-art VQA models suffer from a significant loss of accuracy in this setting.
We also evaluated existing bias-reduction methods.
Even the most general-purpose of these methods do not significantly reduce the use of multimodal shortcuts.
We hope this new evaluation protocol will stimulate the design of better techniques to learn robust VQA models.

\section{Acknowledgements}

The effort from Sorbonne Universit\'e was partly supported by ANR grant VISADEEP (ANR-20-CHIA-0022). This work was granted access to HPC resources of IDRIS under the allocation 2020-AD011011588 by GENCI.

{\small
\bibliographystyle{ieee_fullname}
\bibliography{main}
}



\newpage

\section{Supplementary material}

\subsection{Additional statistics about shortcuts}

\begin{figure}[h!]
    \centering
    \includegraphics[width=.90\linewidth]{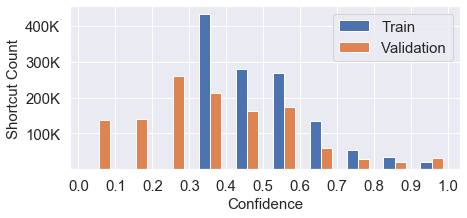}
    \caption{Histogram of shortcuts binned per confidence on the VQA v2 training and validation sets. Our shortcuts are detected on the training set and selected to have a confidence above 30\%.
    Even though their confidence could be expected to be lower on the validation set, it still is above 30\% for a large number of them, indicating that the selection transfers well to the validation set.}
    \label{fig:distribution-rules}
\end{figure}

\paragraph{Confidence distribution on training and unseen data}
Here we show that shortcuts detected on the VQA v2 training set transfer to the validation set.
In Figure~\ref{fig:distribution-rules}, we display the confidence distribution of these shortcuts. As told earlier, we only consider shortcuts that reach a confidence greater than 0.3 on the training set.
The number of shortcuts decreases when the confidence increases.
It is expected to find fewer shortcuts with higher levels of confidence due to the collection procedure of VQA v2 which focused on reducing the amount of data biases and shortcuts.
We evaluate on the validation set the same shortcuts detected on the training set and also display the confidence distribution. We show that our shortcuts are predictive on both training data, and unseen data that follows the training set distribution. The number of shortcuts that reach a confidence between 0.9 and 1.0 is even higher on the validation set than on the training set.
The confidences are overall slightly lower on the validation set, but a large number of them are still above 0.3, indicating that they generalize to new examples from the same distribution.
The great majority of shortcuts, which obtain a confidence lower than 1.0, allows finding examples that contradict them by leading to the wrong answers.
We manually verified by looking at these examples that only a minority are wrongly annotated or ambiguous, most of them are counterexamples.
These counterexamples are the core of our approach to assess the VQA model's reliance on shortcuts.

\paragraph{Distribution of examples per question-type}
 
\begin{figure}[h!]
    \centering
    \includegraphics[width=1.0\linewidth]{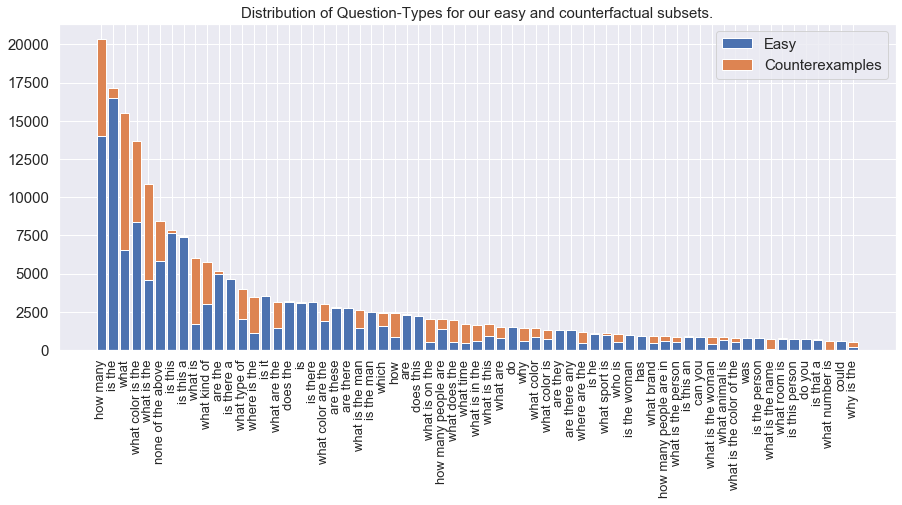}
    \caption{Distribution of the number of examples per question type. Examples associated to our Counterexamples subset are matched by some shortcuts, but no shortcut leads to the correct answer. Examples associated to our Easy subset are matched by at least one shortcut that leads to the correct answer. }
    \label{fig:distrib-question-types}
\end{figure}

In Figure~\ref{fig:distrib-question-types}, we display the distribution of examples per question type, and their split between the Easy and the Counterexamples split.
We show that examples of a question-type that can be answered by \textit{yes} or \textit{no}, such as \textit{is}, \textit{are}, \textit{does}, \textit{do}, mostly belong to the Easy subset. 
Examples of a question-type beginning by \textit{what}, \textit{where} or \textit{why} mostly belong to the Counterexamples subset. These examples need to be answered using a richer vocabulary than \textit{yes} or \textit{no}.
Examples of a question-type beginning by \textit{how} belong to both subset.

\begin{figure}[h!]
    \centering
    \includegraphics[width=1.0\linewidth]{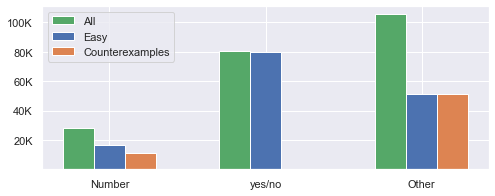}
    \caption{Number of examples per answer type. ``All'' corresponds to all the examples from the VQA v2 validation set. Among them, examples associated to our ``Counterexamples'' subset are matched by some shortcuts, but none of these shortcuts leads to the correct answer. Inversely, examples associated to our Easy subset are matched by at least one shortcut that leads to the correct answer.}
    \label{fig:distrib-answer-types}
\end{figure}

\paragraph{Distribution of examples per answer type}
In Figure~\ref{fig:distrib-answer-types}, we display the distribution of examples in our two subsets per answer type. We see that most yes-no questions are going in the Easy subset, as they are correctly predicted by some rules. On the contrary, for the two other answer types, examples are more evenly distributed between the Easy and Counterexamples subsets.

\subsection{Examples that are not matched by any rule}

\begin{figure}[h!]
    \centering
    \includegraphics[width=1.0\linewidth]{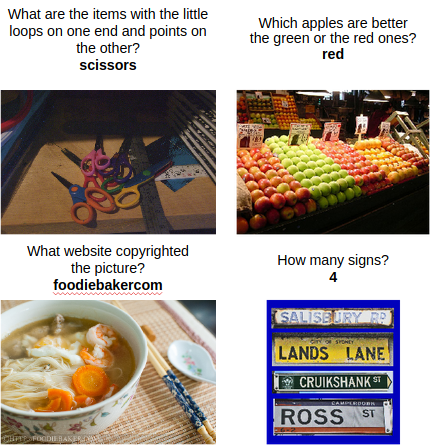}
    \caption{Representative instances of image-question-answer examples that are not matched by any of our shortcuts.
    These examples have unusual questions, images or answers.}
    \label{fig:qual-no-rules}
\end{figure}

In Figure~\ref{fig:qual-no-rules}, we display some representative examples that are neither in the Easy subset nor in the Counterexamples subset. These examples are not matched by any antecedent of our rules. Their input might be unusual. We do not add these examples to our Counterexamples subset, as they do not contradict the shortcuts we found. We discard them entirely from our analysis. There consists in about 3K of examples.

\paragraph{Results with ground-truth visual labels}
We report in Table~\ref{tab:vqa2-results-counterfactual-ground-truth} the results of our analysis with ground-truth visual labels from the COCO~\cite{lin2014microsoftcoco} dataset, instead of labels detected with Faster R-CNN.
We make similar observations to the main experiments of the paper: bias-reductions methods often degrade performances, on both easy and counterexamples split.
A few methods slightly improve the counterexamples score, but much less than on VQA-CP.
The only method which improves both overall  and counterexamples scores is LfF~\cite{nam2020learningfromfailure}. We observed similar results on the dataset with detected labels, reported in Table~1 of the main paper.

\begin{table*}[h!]
    \small
    \centering
    \begin{tabular}{c p{4cm} P{2cm} P{4cm} P{2cm}}
    \toprule
    & Approaches & Overall & \textbf{Counterexamples (ours)} & \textbf{Easy (ours)}\\
    & \multicolumn{1}{r}{\scriptsize \textit{Number of examples}} &  {\scriptsize \textit{214,354}} & {\scriptsize \textit{63,925}} & {\scriptsize \textit{135,324}} \\ 
    \midrule
    {\multirow{3}{*}{\rotatebox[origin=c]{90}{\footnotesize Baselines}}}
    & Shortcuts & 42.14 \white{(+0.00)} & 0.43 \white{(+0.00)} & 65.95 \white{(+0.00)} \\
    & Image-Only & 23.70 \white{(+0.00)} & 2.92 \white{(+0.00)} & 35.39 \white{(+0.00)} \\
    & Question-Only & 44.12 \white{(+0.00)} & 13.98 \white{(+0.00)} & 60.88 \white{(+0.00)} \\
    \midrule
    \multirow{4}{*}{\rotatebox[origin=c]{90}{\footnotesize VQA models}} 
    & SAN~\cite{yang2016stacked} -- \textit{grid features} & 55.61 \white{(+0.00)} & 28.99 \white{(+0.00)} & 70.04 \white{(+0.00)} \\
    & UpDown~\cite{anderson2018bottom} & 63.52 \positive{(+0.00)} & 37.77 \positive{(+0.00)} & 77.52 \positive{(+0.00)} \\
    & BLOCK~\cite{benyounes2019block} & 63.89 \white{(+0.00)} & 37.06 \white{(+0.00)}& 78.52 \white{(+0.00)} \\
    & VilBERT~\cite{lu2019vilbert}  -- \textit{pretrained}$^\dagger$ & 67.77 \white{(+0.00)} & 43.32 \white{(+0.00)} & 81.27 \white{(+0.00)} \\
    \midrule
    \multirow{8}{*}{\rotatebox[origin=c]{90}{{\footnotesize Bias-reduction methods}}} 
    & \multicolumn{4}{c}{\textit{UpDown~\cite{anderson2018bottom} is used as a base architecture for bias-reduction methods}} \\
&  RUBi~\cite{cadene2019rubi}  & 61.88 \negative{(-1.64)} & 36.05 \negative{(-1.72)} & 75.84 \negative{(-1.68)} \\
 &  LMH + RMFE~\cite{gat2020rmfe}  & 60.12 \negative{(-3.40)} & 34.97 \negative{(-2.80)} & 73.80 \negative{(-3.72)} \\
 &  ESR~\cite{shrestha2020negativecase}  & 62.96 \negative{(-0.56)} & 37.22 \negative{(-0.55)} & 76.98 \negative{(-0.54)} \\
 &  LMH~\cite{clark2019ensemblebias}  & 61.15 \negative{(-2.37)} & 37.82 \positive{(+0.05)} & 73.91 \negative{(-3.61)} \\
 &  LfF~\cite{nam2020learningfromfailure}  & 63.57 \positive{(+0.05)} & 38.18 \positive{(+0.41)} & 77.44 \negative{(-0.08)} \\
 &  LMH+CSS~\cite{chen2020counterfactual}  & 53.55 \negative{(-9.97)} & 37.27 \negative{(-0.50)} & 62.30 \negative{(-15.22)} \\
 &  RandImg~\cite{teney2020value}  & 63.34 \negative{(-0.18)} & 38.13 \positive{(+0.36)} & 77.05 \negative{(-0.47)} \\
    \bottomrule \\
    \end{tabular}
    \caption{Results of our VQA-CE evaluation protocol with \textbf{ground-truth visual labels}. We report accuracies on VQA v2 full validation set and on our two subsets: \textbf{Counterexamples} and \textbf{Easy} examples. We re-implemented all models and bias-reduction methods.
    $^\dagger$VilBERT is pretrained on Conceptual Caption and fine-tuned on VQA v2 training set. Scores in {\positive{(green)}} and {\negative{(red)}} are relative to UpDown~\cite{anderson2018bottom}.
    }
    \label{tab:vqa2-results-counterfactual-ground-truth}
\end{table*}


\paragraph{Results on VQA v1}

We report in Table~\ref{tab:vqa1-results-counterfactual} the results of our analysis on the VQA v1 dataset. We observe similar results as in Table~1 from the main paper. Most bias-reduction methods degrade performances on the counterexamples split, and only LfF~\cite{nam2020learningfromfailure} improves performances on all three splits.

\begin{table*}[h!]
    \small
    \centering
    \begin{tabular}{c p{4cm} P{2cm} P{4cm} P{2cm}}
    \toprule
    & Approaches & Overall & \textbf{Counterexamples (ours)} & \textbf{Easy (ours)}\\
    & \multicolumn{1}{r}{\scriptsize \textit{Number of examples}} &  {\scriptsize \textit{121,512}} & {\scriptsize \textit{40,052}} & {\scriptsize \textit{80,539}} \\ 
    \midrule
    {\multirow{3}{*}{\rotatebox[origin=c]{90}{\scriptsize Baselines}}}
    & Shortcuts &  44.71 \white{(+0.00)} & 0.05 \white{(+0.00)} & 67.35 \white{(+0.00)} \\
    & Image-Only & 24.39 \white{(+0.00)} & 1.75 \white{(+0.00)} & 35.83 \white{(+0.00)} \\
    & Question-Only & 49.20 \white{(+0.00)} & 13.48 \white{(+0.00)} & 67.27 \white{(+0.00)} \\
    \midrule
    & SAN~\cite{yang2016stacked} -- \textit{grid features} & 58.35 \white{(+0.00)} & 26.09 \white{(+0.00)} & 74.58 \white{(+0.00)} \\
    & UpDown~\cite{anderson2018bottom} & 62.83 \positive{(+0.00)} & 31.71  \positive{(+0.00)} & 78.49 \positive{(+0.00)} \\
    \midrule
    \multirow{7}{*}{\rotatebox[origin=c]{90}{{\scriptsize Bias-reduction methods}}} 
    & \multicolumn{4}{c}{\textit{UpDown~\cite{anderson2018bottom} is used as a base architecture for bias-reduction methods}} \\
    &  RUBi~\cite{cadene2019rubi} & 55.82 \negative{(-7.01)} & 23.87 \negative{(-7.84)} & 71.90 \negative{(-6.59)} \\
     &  LMH + RMFE~\cite{gat2020rmfe} & 62.97 \positive{(+0.14)} & 31.09 \negative{(-0.62)} & 79.02 \positive{(+0.53)} \\
     &  ESR~\cite{shrestha2020negativecase}  & 63.03 \positive{(+0.20)} & 31.50 \negative{(-0.21)} & 78.91 \positive{(+0.42)} \\
     &  LMH~\cite{clark2019ensemblebias}  & 59.74 \negative{(-3.09)} & 32.80 \positive{(+1.09)} & 73.30 \negative{(-5.19)} \\
     &  LfF~\cite{nam2020learningfromfailure}  & 63.26 \positive{(+0.43)} & 32.05 \positive{(+0.34)} & 78.97 \positive{(+0.48)} \\
     &  RandImg~\cite{teney2020value}  & 62.87 \positive{(+0.04)} & 31.09 \negative{(-0.62)} & 78.87 \positive{(+0.38)} \\
     
    \bottomrule \\
    \end{tabular}
    \caption{Results of our VQA-CE evaluation protocol on \textbf{VQA v1} full validation set and on our two subsets: \textbf{Counterexamples} and \textbf{Easy} examples. We re-implemented all models and bias-reduction methods.
    Scores in {\positive{(green)}} and {\negative{(red)}} are relative to UpDown~\cite{anderson2018bottom}.
   }
    \label{tab:vqa1-results-counterfactual}
\end{table*}


\newpage
\subsection{Rules with supporting examples and counterexamples}

\begin{figure*}[h!]
    \centering
    \includegraphics[width=0.9\linewidth]{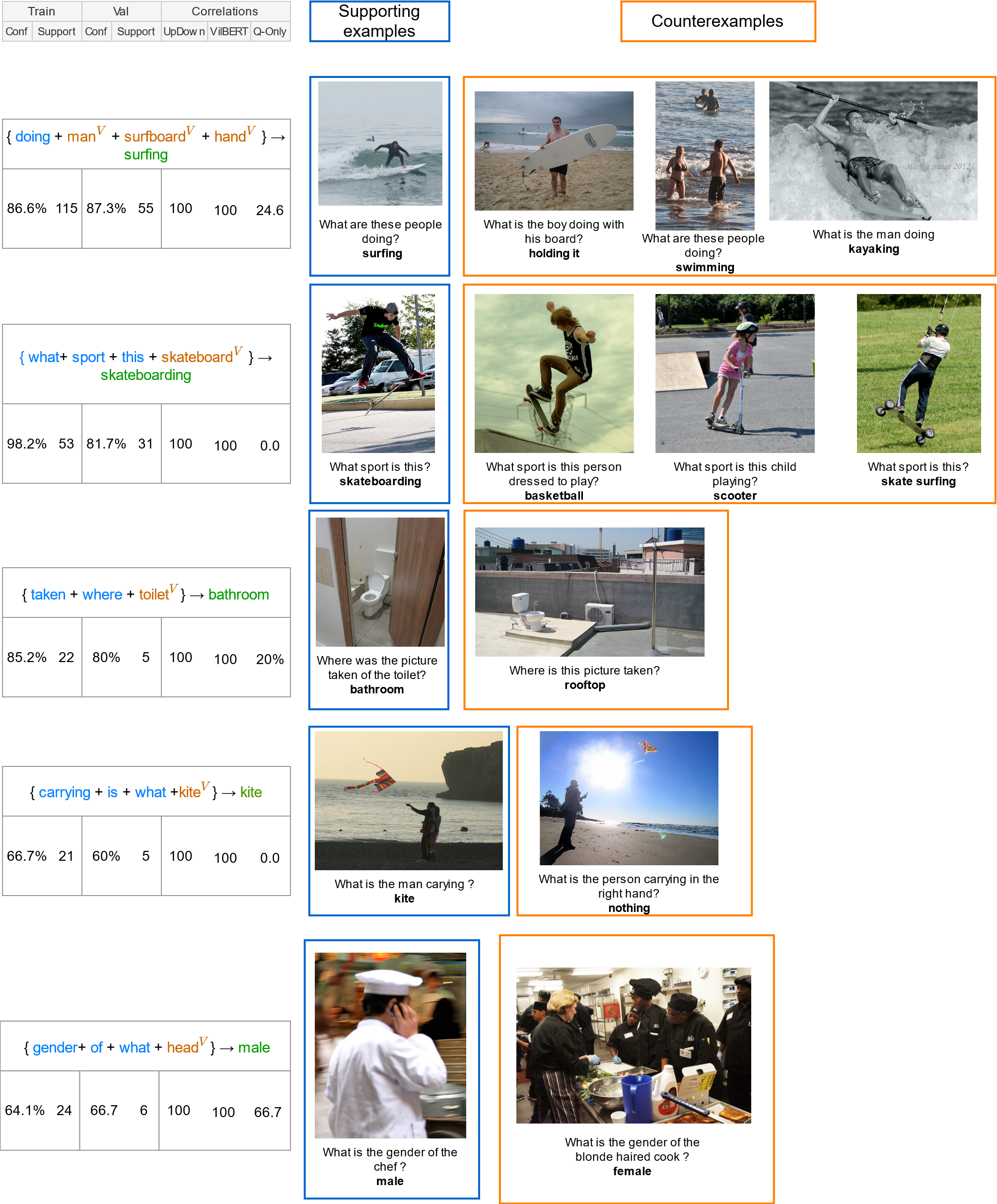}
    \caption{Instances of shortcuts that are highly correlated with VQA models' predictions. We display their antecedent made of \textual{words} from the question and \visual{objects} from the image, and their \answer{answer}. Their support, i.e. number of examples matched by the antecedent, and confidence, i.e. percentage of correct answers among them, have been calculated on the VQA v2 training and validation sets. We report the correlation coefficients of their predictions with those of three VQA models: UpDown~[3] that uses an object detector, VilBERT~[31] that has been pretrained on a large dataset, and Q-only~[21] that only uses the question. We also display some supporting examples, in blue, and counterexamples, in orange.}
    \label{fig:qual-counterexamples}
\end{figure*}

In Figure~\ref{fig:qual-counterexamples}, we display some counterexamples to some rules displayed in Table~2 of the main paper. Some of those examples are ``true'' counterexamples, where the input does match the rule's antecedent, but the answer is different. For instance, in the first example of the first rule, the question is actually about the clothes and not the sport, and the man is dressed in a basketball outfit. 
On the contrary, some examples are there due to an incorrect object detection: in the second example of the first rule, the object detection module detected a skateboard instead of a scooter. Thus, the example is incorrectly matched. 

\end{document}